# Keypoint Autoencoders: Learning Interest Points of Semantics

Ruoxi Shi, Zhengrong Xue, and Xinyang Li

*Shanghai Jiao Tong University*

**Abstract**

Understanding point clouds is of great importance. Many previous methods focus on detecting salient keypoints to identity structures of point clouds. However, existing methods neglect the semantics of points selected, leading to poor performance on downstream tasks. In this paper, we propose Keypoint Autoencoder, an unsupervised learning method for detecting keypoints. We encourage selecting sparse semantic keypoints by enforcing the reconstruction from keypoints to the original point cloud. To make sparse keypoint selection differentiable, Soft Keypoint Proposal is adopted by calculating weighted averages among input points. A downstream task of classifying shape with sparse keypoints is conducted to demonstrate the distinctiveness of our selected keypoints. Semantic Accuracy and Semantic Richness are proposed and our method gives competitive or even better performance than state of the arts on these two metrics.

**Keywords:** 3D Keypoint Detecting, Machine Vision, Deep Learning

## 1. Introduction

Point cloud is considered to be an irregular data format [Qi et al., 2017]. Due to the large number of data points, processing a point cloud is often challenging. One way to handle this problem is to represent a large-scale dense point cloud with a relatively small-scale sparse keypoints. Traditional geometric based keypoint detectors like [Harris and Stephens, 1988] fail to extract semantic-rich information from an arbitrary object and their success heavily depends on their manual parameter selections.

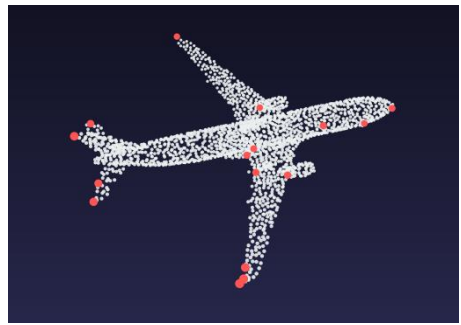

Figure 1: **Example of Detected Keypoints.**

In the trend of deep learning, Qi et al. proposed PointNet [Qi et al., 2017], a novel neural network capable of supervised learning tasks on point clouds. Unfortunately, as the understanding towards which keypoints are of semantics may vary in person, it is virtually impossible to obtain ground-truth keypoint labels. As a result, few supervised data-driven methods prove to be successful in selecting keypoints from a point cloud. Recently, some unsupervised methods [Li and Lee, 2019][Yew and Lee, 2018] are proposed. However, they only stress high repeatability of keypoints and neglect semantics, thus behaving poorly for downstream tasks especially under small number of keypoints.

In view of the challenges above, we propose Keypoint Autoencoder (KAE). Autoencoders have long been utilized to reduce redundant information in signals. Similarly, we utilize them to reduce 'redundant' points in the point cloud to detect keypoints. We slightly modified the traditional autoencoder so that a set of keypoints is used as the latent representation instead of an arbitrary latent vector. To this end, the encoder outputs a set of probability distributions on input points, and a proposal module is invoked on the distributions to get the final keypoints. The decoder tries to reconstruct the original point cloud from the keypoints. To make the hard keypoint selection process differentiable, Soft Keypoint Proposal is proposed, the main idea of which is to compute an average among input points weighted by the probability distributions. Furthermore, on the basis of KAE, Auxiliary Classifier Keypoint Autoencoder (AC-KAE) is built where class information of the point cloud is added to assist downstream tasks and class-wise feature learning. In our framework, semantic information in detected keypoints is greatly encouraged since it is impossible to reconstruct each object with only shape information and no semantics, especially under low quantities of keypoints demanded.

We verify the performance of our method with two different metrics: distinctiveness in terms of downstream classifier accuracy and semantic information quality. We propose **Semantic Accuracy** and **Semantic Richness** to comprehensively decide the quality of semantic information expressed by the keypoints. A Mean Opinion Score test is conducted using these two indexes. An example of keypoints detected with our method is shown in figure 1. It can be seen that points with semantics (head and tail, roots of engines, tip of wings) are detected as keypoints.

## 2. Related Work

PointNet [Qi et al., 2017] and its variant PointNet++ [Qi et al., 2017] exploit max pooling, a symmetric function, to process unordered, varying-length point clouds into a regular, fixed-length global feature vector, which enables deep learning networks to consume point clouds. They and many other PointNet-based networks are proposed to deal with various 3D vision tasks such as classification, segmentation and pose estimation. However, there are still few unsupervised methods designed for extracting keypoints from a point cloud. So far, USIP [Li and Lee, 2019] is considered to be the state-of-the-art method. USIP takes advantage of probabilistic chamfer loss to obtain highly repeatable keypoints. However, keypoints generated from USIP don't necessarily contain much semantic information, which makes USIP less helpful for actual downstream tasks. Meanwhile, S-NET [Dovrat et al., 2019] is aware of downstream tasks, which certainly avoids the demerits of USIP. However, the nature that it is tailored to specific tasks disqualifies S-NET to be a keypoint detector, as the point set it selects changes w.r.t. different downstream tasks. In comparison, our method takes semantics into consideration, which is helpful for downstream tasks. Furthermore, the point set we select only depends on the raw point cloud, and thus it is invariant to specific tasks.

## 3 Method

**The Pipeline.** There are four major modules in our design: an encoder, a decoder, a proposal module to obtain keypoints from the encoder, and an optional auxiliary classifier. The architecture is shown in figure 2.

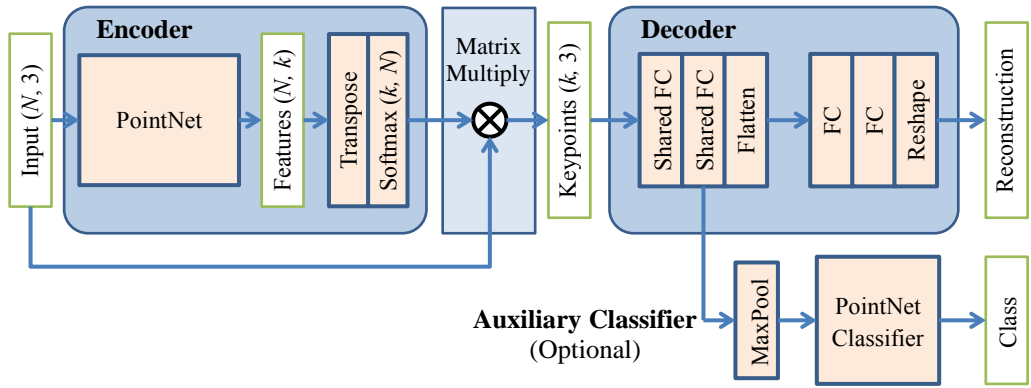

Figure 2: **Architecture of Keypoint Autoencoder.**

**Encoder.** Suppose we want the system to extract $k$ keypoints. Then the encoder inputs a point cloud (shape $[N, 3]$), which goes through a PointNet [Qi et al., 2017] to obtain point-wise features (shape $[N, k]$). After that it is transposed and a softmax activation is applied to obtain $k$ probability distributions (shape $[k, N]$).

**Soft Keypoint Proposal.** This small module is invoked on the distributions to get $k$ keypoints (shape $[k, 3]$). These points act as the input to decoder. However, no gradient would be propagated into the encoder if keypoints were directly sampled from input. Upon this we propose the Soft Keypoint Proposal: Input points are weighted by the distributions obtained from the encoder, the weighted average of which is passed into the decoder. Formally, if we denote $K_s(i)$ as the soft keypoints, $X_j$ as the $j$-th input point (each a 3-D vector), and $D_{k \times N}$ is the outputs of the encoder (after applying softmax), then

$$K_s(i) = \sum_{j=1}^{N} D_{ij} X_j, i = 1, 2, 3, ..., k$$

Note that it is equivalent to a matrix multiplication as shown in figure 2.

In the detection process, an ordinary selection operation is performed according to the probability distributions obtained from the encoder. Points with highest probabilities are selected with NMS (Non Maximal Suppression) until the number of points selected reaches the requested value. These points act as the final keypoints detected.

**Decoder.** The decoder is composed of fully connected (FC) layers. The first block shares weights among points to get point-wise features (shape $[k, 64]$). The features are reconstructed into a point cloud of same size as the input point cloud (shape $[N, 3]$) by the second block. Chamfer loss [Barrow and Tenenbaum, 1977] is applied between the reconstructed point cloud and the original one:

$$L_c = \sum_{x \in S_1} \min_{y \in S_2} \|x - y\|_2^2 + \sum_{y \in S_2} \min_{x \in S_1} \|x - y\|_2^2$$

Where $S_1, S_2 \subset R^3$ are original and reconstructed point clouds, respectively.

**Auxiliary Classification.** In addition to the point cloud itself, meta information such as the class of the object that yields the point cloud is usually available in the dataset, which may help improve the distinctiveness of selected keypoints. Some common features shared by a whole class of point clouds such as symmetries and object structures can be learned to improve semantics of the keypoints. Thus we propose Auxiliary Classifier Keypoint Autoencoder (AC-KAE). A PointNet classifier branches from the keypoint feature layer and contributes gradient from classification results to the network. The branch is marked as optional in the figure.

## 4. Experiment

In this section, we demonstrate that our method shows strong performance detecting keypoints with semantic information of high distinctiveness or is even better than state of the art.

### 4.1. Distinctiveness: Downstream Classifier Accuracy

Distinctiveness is an important metric of the keypoint detector. A set of distinctive keypoints should describe an object uniquely and one should be able to clearly classify an object based on these sparse keypoints. Therefore, we come up with a downstream task described below.

For each point cloud in the ModelNet40 [Wu et al., 2015] dataset, different detectors are applied to obtain 8 keypoints. With only these keypoints as input, a same PointNet classifier is trained. The overall accuracy is evaluated after 50 epochs. For the KAE based models we proposed, the soft keypoints are adopted.

|          | FPS   | S-NET | USIP  | KAE (ours) | AC-KAE (ours) |
|----------|-------|-------|-------|------------|---------------|
| Accuracy | 68.3% | 83.6% | 56.8% | 83.7%      | **85.7%**     |

Table 1: **Accuracy of Downstream PointNet Classifier on the ModelNet40 Classification Problem.**

Our methods achieve better results than previous models, which means that our autoencoder based detectors are better in distinctiveness. The USIP detector trains against repeatability as [Dovrat et al., 2019] stated in the paper, and it can be concluded that the USIP detector loses distinctiveness with a small quantity of requested keypoints.

### 4.2. Semantic Information

Semantic information is a vital property of interest points. We propose two quantities, namely **Semantic Accuracy** and **Semantic Richness** to describe the accuracy and completeness of semantic information expressed with the keypoints selected.

It is hard to formally define these metrics. Basically, high **Semantic Accuracy** is achieved when the points selected are those of semantics in the point cloud. High **Semantic Richness** is achieved when most of the semantics are covered. However, the relation between computable metrics and these two quantities is unclear and may depend on specific point cloud structure. As a result the metrics can only be subjectively compared among different keypoint detectors. Here some examples are given in figure 3 together with a basic analysis on the semantic information of selected keypoints from the three models.

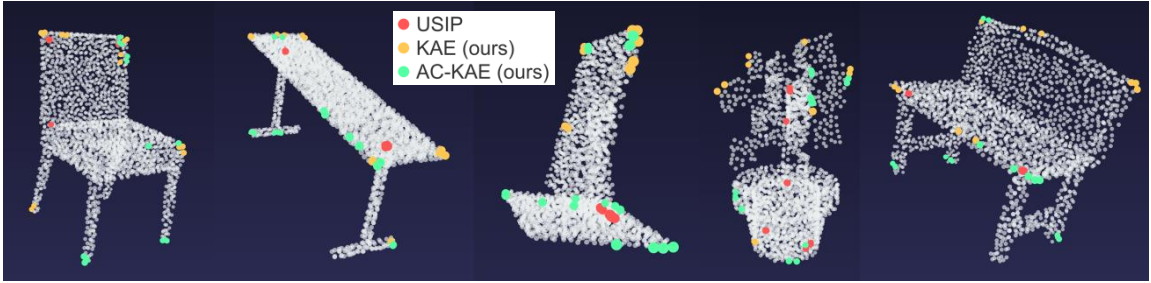

Figure 3: **Examples of Keypoints Detected by USIP, KAE and AC-KAE Detectors.** It can be seen that the keypoints from USIP mainly fall on joints near the central area, making their semantic information accurate but not rich. The keypoints from our methods are mainly at corners of the cloud, but also appear on some major joints. It also demonstrates AC-KAE model's interesting ability to make use of the symmetries from classes of the point cloud, thus sparing more keypoints to represent extra semantic information.

**Mean Opinion Score (MOS) Test.** To further evaluate the semantic information contained in the keypoints generated by our proposed models, we conducted an MOS test among 28 undergraduate students and engineers majoring in relative fields. Detectors are used to extract 16 keypoints from 10 randomly chosen point clouds from the ModelNet40 [Wu et al., 2015] dataset. After viewing the keypoints together with the original point cloud, the subjects are asked to grade the semantic accuracy and richness of keypoints in a 10-point scale score. To ensure the effectiveness of the test, the subjects are all blind to the labels.

|  | USIP | KAE (ours) | AC-KAE (ours) |
|---|---|---|---|
| Semantic Accuracy | 4.571 | **6.049** | 5.546 |
| Semantic Richness | 4.157 | **5.814** | 5.556 |

Table 2: **Subjective 10-scale Mean Opinion Scores.**

It can be seen from the table that our KAE method achieves highest scores, and is significantly better than USIP. However, the AC-KAE model performs worse than the KAE model. A possible cause is that our AC-KAE method makes use of the symmetries and avoids points with similar semantics, so some parts of point clouds are neglected, leading to a drop in subjective opinion scores.